# BioSimplify: an open source sentence simplification engine to improve recall in automatic biomedical information extraction

Siddhartha Jonnalagadda, Graciela Gonzalez
Department of Biomedical Informatics, Arizona State University, Phoenix, AZ

**Abstract**

*BioSimplify is an open source tool written in Java that introduces and facilitates the use of a novel model for sentence simplification tuned for automatic discourse analysis and information extraction (as opposed to sentence simplification for improving human readability). The model is based on a "shotgun" approach that produces many different (simpler) versions of the original sentence by combining variants of its constituent elements. This tool is optimized for processing biomedical scientific literature such as the abstracts indexed in PubMed. We tested our tool on its impact to the task of PPI extraction and it improved the f-score of the PPI tool by around 7%, with an improvement in recall of around 20%. The BioSimplify tool and test corpus can be downloaded from https://biosimplify.sourceforge.net*

**Introduction**

Explaining the limitation of the bag of words model, linguist Zellig Harris pointed out[1]: "language is not merely a bag of words but a tool with particular properties ... The linguist's work is precisely to discover these properties". When discourse analysts try to discover these properties, they usually break the sentence into simpler clauses. It should be noted however that a single independent clause or simple sentence as defined by Quirk[2], may still be complex. Consider for example the following sentence, which even if one were to break it into simpler clauses, it would still be a complex: *We have identified a new TNF-related ligand, designated human GITR ligand (hGITRL), and its human receptor (hGITR), an ortholog of the recently discovered murine glucocorticoid-induced TNFR-related (mGITR) protein.* Gee[3] advises that critically analyzing the discourse involves separating and unpacking clauses from sentences and phrases to understand all the perspectives. Automatically creating this set of simplified sentences for the purpose of information extraction on biomedical text is the subject of this paper. While existing NLP methods for information extraction already use grammatical information of text in the form of features like POS tags, parse trees and dependencies – informally known as "bag of NLP", the usual focus is to choose one optimal such parsing for further processing. Our approach is rather a "shotgun" approach: use grammatical information in elemental chunks that can then be combined and recombined to generate many sentences from one (different perspectives) in order to maximize the likelihood that an automatic extraction engine can find in one (or several) of them the information contained in the original sentence. Thus, BioSimplify outputs the set of all sentences it can generate from the original sentence such that they are: 1) implied by the original sentence, 2) grammatically correct, and 3) shorter than the original sentence.

**Background**

*Sentence Simplification.* To understand the value of this approach, we need to recognize the different motivations for sentence simplification. One of the applications of sentence simplification is to create sentences that are more readable to humans. We see examples of this in the works of Siddharthan[4] and Carroll[5]. These projects aim to create sentences that are shorter, grammatically correct, information-preserving and cohesive (the property where the context of a discourse element can be inferred from its precedents). Sentence simplification is also applied in text summarization systems like SumBasic[6] where the focus is to preserve only the important content. Such an approach called sentence shortening or sentence compression doesn't necessarily preserve semantic content. Our goal is rather to use sentence simplification for improving the automatic extraction process. We have previously shown that simplification improved the performance of parsers[7]. There the goal was to preserve both semantic content and grammatical correctness, but not necessarily cohesiveness. Our present work shares the same properties, but is based on a different approach that lends itself better for open-source publication of the engine, scalability, and generalization to information extraction tasks (particularly named entity recognition and association extraction) and other NLP applications, like semantic role labeling.

*Protein-Protein Extraction.* The study of protein-protein interactions and other molecular events is a central tenet of modern translational and genomic research. Publications centering on reports of such atomic events abound, and their manual extraction from the literature currently occupies many trained curators that deposit them in databases such as DIP, MINT, or IntAct. Manual curation, however, despite years of effort, has only made a small dent (calculated at around 7%) into the volume of publications believed to report protein-protein interactions. Automatic extraction of such facts is thus a priority for biomedical text mining researchers, although performance is still poor. We attempt to increase the performance of extraction systems by reducing the complexity of sentences that could be hiding PPIs. In the recent BioCreative II.5 international competition for extracting protein-protein interactions, our team[9] introduced the preprocessing step of replacing noun phrases in the sentences with the head noun, a technique that helped us achieve the best F-score for the task. This paper presents a tool complete in itself for sentence simplification.

**Methods**

While trying to understand how to break a sentence into simpler parts, we focused on how sentences grow, and devised methods to undo the expansion. Halliday[8] states that there are three ways for such expansion: 1) elaborating an existing basic structure, 2) extending it by addition or replacement, and 3) enhancing its environment. We used these basic guidelines to design rules (listed in Table 2) for creating simpler sentences out of a complex one. The rest of this section presents the Noun Phrase Replacement module and the Syntactic Simplification algorithm along with the rules (Tables 1 and 2).

**Table 1.** Syntactic Simplification algorithm.

---

synSimp(t), where t is the Penn tree of the given sentence:

1) Initialize <u>simpTrees</u>, the ordered set containing the Penn trees of all simplified sentences, with the Penn tree of the original sentence

2) FOR EACH subtree of t traversed in the order of depth-first traversal
 - perform necessary simplifications at that node which are the simplifications that needn't be repeated for all the parents to this node

3) Add the present tree to <u>simpTrees</u>

4) FOR EACH unprocessed tree in <u>simpTrees</u>
- FOR EACH subtree of t traversed in the order of depth-first traversal
- perform the simplifications for this node
- add new trees in <u>simpTrees</u> if applicable

5) return the sentences represented by the trees in <u>simpTrees</u>

---

**1. Noun Phrase Replacement**

A noun phrase in English consists of an optional determinative, an optional premodifier, a mandatory head noun, and an optional postmodifier[2]. Noun Phrase chunkers usually return the noun phrases of the smallest length, excluding postmodifiers. Hence, the last word of an identified noun phrase is the head noun. Previously[9], we introduced the preprocessing step of replacing noun phrases in the sentences with their head noun. However, removal of the optional determinative makes the sentence grammatically incorrect, while removal of the premodifier still gives a grammatically correct sentence. So, in this work, we only remove the premodifiers. For example, the noun phrase "*the recently discovered murine glucocorticoid*" is replaced with "*the glucocorticoid*". The part-of-speech (POS) tags in the sentence are identified using Lingpipe[1], one of the most widely used POS taggers trained on GENIA biomedical corpus[2]. We then use the OpenNLP maximum entropy method[3] to identify the noun phrases in the sentence. The other chunkers that could be considered are GATE chunker, GENIA Tagger, Lingpipe and Yamacha. For noun phrase chunking, GENIA Tagger and OpenNLP perform the best (f-score of 90% on GENIA Corpus), but OpenNLP is more usable in our system workflow as it is written in Java, while the GENIA Tagger is written in C++. To remove the premodifiers, all the tokens other than the head noun and the starting determinative or numeral (if they exist) are removed from the noun phrases.

**2. Syntactic Simplification**

While POS taggers only indicate the grammatical role of a particular word, a parse tree represents the syntactic structure of the whole sentence, giving complete details on how the words in it are related to each other. Sentence simplification systems[6,10,11] usually have parsers as integral part of their algorithm, while there are few systems[4] that use only POS. The latter aim for fast simplification at the point of application, while the former give a higher importance to accuracy of the output. BioSimplify is optimized for accurate and flexible biomedical information extraction, and thus uses the output from parsers. This is a reasonable choice considering that there has been significant increase in the accuracy of parsing biomedical text from 80% in 2005[12], to 84% in 2008[13], and to 88% in 2009[4] measured according to f-score and it facilitates inclusion of the simplification algorithm in different NLP pipelines. This choice also decouples the simplification process from the parsing step, allowing it to be done separately or to take advantage of the availability of collections of pre-processed sentences like the NLP web service provided by NCIBI[14] or the PTDB[15] database of Arizona State University. We currently use Penn trees obtained from McClosky parser[13] which has an f-score of 88% for parsing biomedical text. BioSimplify can also be used with Penn trees produced from other parsers like Stanford parser and Link Grammar that can produce PTB-style[16] output and also from Penn trees created apriori. Our goal is to produce all possible grammatically correct simplified sentences, assuming the available Penn tree is completely accurate. Table 1 describes the algorithm for syntactic simplification to produce grammatically correct sentences.

---

[1] http://alias-i.com/lingpipe
[2] http://www-tsujii.is.s.u-tokyo.ac.jp/~genia/topics/Corpus/
[3] http://maxent.sourceforge.net
[4] http://www.cs.brown.edu/~dmcc/biomedical.html

**Table 2:** Syntactic simplification rules with explanation and example. Column 1 is the mathematical notation for the rule and Column 2 is the verbal explanation of it. Readers interested in adapting the rules in their own environment might be more interested in Column1 than others.

| Rule | Explanation | Example | Result |
|---|---|---|---|
| S ~ {S}. S contains NP S contains VP | Adds all simple sentences (phrases present in a larger sentence, but in themselves are a grammatical sentence) | In differentiating C2C12 cells, E2F complexes switch and DNA synthesis in response to serum are prevented when MyoD DNA binding activity and the cdks inhibitor MyoD downstream effector p21 are induced. | MyoD DNA binding activity and the cdks inhibitor MyoD downstream effector p21 are induced. |
| NP[NP1 VP1*] ~ [NP1] {NP1 "can be" VP1} *VP1 starts with a gerund, present participle or past participle | NP Postmodification by verb phrase (separate the verb phrase that provides extra information about the noun phrase, but is not related to the whole sentence) | The cloning of members of these gene families and the identification of the protein-interaction motifs found within their gene products has initiated the molecular identity of factors (TRADD, FADD/MORT, RIP, FLICE/MACH, and TRAFs) associated with both of the p60 and p80 forms of the TNF receptor and with other members of the TNF receptor superfamily. | The cloning of members of these gene families and the identification of the protein-interaction motifs has initiated the molecular identity of factors… The protein-interaction motifs can be found within their gene products. |
| NP[NP1 ADJP1] ~ [NP1] {NP1 "can be" ADJP1 } | NP Postmodification by adjective phrase (similar to above) | Src homology domain-2 (SH2)/SH3 domain - can be containing adapters such as Grb2, Crk, and Crk-L, which interact with guanine nucleotide exchange factors specific for the Ras family. | … interact with guanine nucleotide exchange factors. Guanine nucleotide exchange factors can be specific for the Ras family. |
| NP[NP1 PP] ~ [NP1] | NP Postmodification by prepositional phrase is removed | To explore the role of the different domains of the betaL subunit in IFNalpha signaling, we coexpressed wild-type alpha subunit and truncated forms of the betaL chain in L-929 cells. | To explore the role in IFNalpha signaling, we coexpressed … |
| VP[MD VP1 , S*] ~ [MD VP1] *S contains VP and not NP | VP Postmodification by verb phrase is removed | T lymphocytes can be activated normally in response to either stimulus, demonstrating that the effects of the inactive CaMKIV on activation are reversible. | T lymphocytes can be activated normally in response to either stimulus. |
| VP[… , PP] ~ }, PP{* *Terminal prepositional phrase and preceding comma are removed from verb phrase | VP Postmodification by prepositional phrase is removed | Because cell lines can lose their differentiated phenotype in culture across passages, documentation of gene expression must be determined for passage populations, for us to have knowledge of cell behavior in vitro. | Because cell lines can lose their differentiated phenotype in culture across passages, documentation of gene expression must be determined for passage populations. |
| NP[NP1 PRN] ~ [NP1] [PRN - LRB - RRB]* *The left and right brackets are removed from the parenthetical | Handling abbreviations: Replace with two sentences- one with abbreviation removed, the other with NP replaced by abbreviation | Coexpression of the alpha and betaL subunits of the human interferon alpha (IFNalpha) receptor is required for the induction of an antiviral state by human IFNalpha. | Coexpression of the alpha and betaL subunits of the human interferon alpha receptor is … Coexpression of the alpha and betaL subunits of the human IFNalpha is … |
| NP[NP : S*] ~ [S*] *S contains VP or NP | Section indicator is removed | OBJECTIVE: To investigate the relationship between the expression of Th1/Th2 type cytokines and the effect of interferon-alpha therapy. | To investigate the relationship between the expression of Th1/… |
| S[S1* , NP VP] ~ [NP VP] *S1 doesn't contain both NP and VP | Content clause is removed | To characterize these pathways, we focused on changes in the cyclin-dependent kinase inhibitors and their binding partners that underlie the cell cycle arrest at senescence. | We focused on changes in the cyclin-dependent kinase inhibitors and their … |
| NP[NP SBAR] ~ [NP], {SBAR - WHNP + NP}* *Wh-NP in the relative clause is replaced by NP from main clause | Relative Clause of this type is separated from the original sentence | To characterize these pathways, we focused on changes in the cyclin-dependent kinase inhibitors and their binding partners that underlie the cell cycle arrest at senescence. | …changes in the cyclin-dependent kinase inhibitors and their binding partners. The cyclin-dependent kinase inhibitors and their binding partners underlie… |
| VP[…, SBAR…] ~ }, SBAR{ | Relative Clause of this type is removed | As [Ca2+]o increased, [Ca2+]i rapidly increased, as monitored by fluorometry. | As [Ca2+]o increased, [Ca2+]i rapidly increased. |
| VP , CC VP2] ~ [VP1] [VP2] PP [PP1, CC PP2] ~ [PP1] [PP2] ADJP[ADJP1 , CC ADJP2] ~ [ADJP1] [ADJP2] | Coordination of verb phrases, or prepositional phrase or adjective phrase is removed by separating the coordinates into different sentences | These mechanisms must be understood in order to prevent, or combat, the emergence of a virulent, multidrug-resistant form of the bacillus that would be uncontrollable by means of today's treatment strategies. | These mechanisms must be understood in order to prevent, the emergence of a virulent, multidrug … These mechanisms must be understood in order to combat , the … |

The algorithm has a time complexity of $O(n^2*R)$, where n is the number of tokens in the sentence and R is the number of simplifications rules. The average time complexity is, however, $O(nlog(n)*R)$. One of the features of BioSimplify is avoidance of domain-specific rules. For example, we don't replace entity names (like genes) with shorter alternatives as is done with the noun phrases in the present version and with the gene names in our earlier version[11]. We also avoided hard-coding the words in the rules created to split sentences with relative clauses. These measures enhance the domain adaptability of the system. Table 2 explains some of the simplification rules we used. There are around 40 syntactic simplification rules available with the source code and documentation. The format for the rules is A[A1 A2 … An] ~ [B1...Bp] {C1…Cq} {D1…Dr} , where the variables are non-terminals in the parse tree. [A1 A2 … An] are the children of A in the original tree. [B1...Bp] are preserved from the original tree, {C1…Cq} are removed from the original tree and added as a separate tree, and {D1…Dr} are removed from the original tree. To make sure that the rules are optimized for biomedical sentences, we manually examined each sentence in GENIA biomedical corpus and designed rules that would create the largest possible "bag of simplified sentences" based on Halliday's formalism for sentence simplification. Some rules are classified as necessary (Table 2), and are executed only once since that transformations based on those rules do not decrease the semantic content of the original sentence. However, some transformations, like the post-modification of noun phrases by prepositional phrases, often destroy the semantic content. For such transformations, the algorithm ensures that both the simplified sentences and the sentence from which they are derived are preserved. Our rule-set consists of all the rules present in the most comprehensive rule-based system currently known[4] – albeit using a different notation based on phrase structures – and includes many additional rules. Siddharthan's system[4] handles coordination only at the clause level, while we handle it also at the phrase level. We also remove section indicators, content clauses, and post-modifiers of phrases not handled before. Premodification of noun phrases is handled at NP replacement stage. We currently don't handle pronoun resolution.

**PPI Extraction Evaluation**
For the purpose of evaluating the impact of sentence simplification, we use AIMed[17] corpus (which is extensively used in comparing PPI extraction methods) and PIE[18] (a machine-learning based approach available as a web service that uses the parse tree information from the Collins statistical parser as its key component). PIE returns two kinds of results – one with a high precision, which we call *tight PIE*; and the other with low precision, which we call *light PIE*. We also compare the present version of BioSimplify with the older version[11] which is limited in its functionality because it only implements the rules described by Siddharthan[4]. The present version which has an average time complexity of $O(nlog(n)*R)$ is faster than the older version which has a time complexity of $O(n^3*R)$, where n is the number of tokens in the sentence and R is the number of rules. The older version has domain specific optimizations (like replacing the gene names with single-word identifiers), which were not used in the newer version for portability.

**Results and Conclusion**
We used PIE to test for the presence of PPIs before and after simplification in both the versions. The test set is from 18 PubMed abstracts in AIMed with ids between 9121766 and 9427624. Overall, out of the 189 sentences in these abstracts, 63 contain PPI(s). The aggregate results of PIE are presented in Tables 3 and 4. Precision, Recall, and F-score assume conventional meaning. Using BioSimplify improved the performance of light PIE by 9% in f-score and in recall by 24%. It also enabled an improvement in f-score by 7% and in recall by 20% on tight PIE. These improvements are statistically significant based on the two-tailed paired t-test on the outputs (PPI present/absent) of PIE before and after simplification ($p=1.3 \times 10^{-5}$ for light PIE and $p=7.2 \times 10^{-8}$). Overall, the present version of BioSimplify performs much better than the older version. This is because using the "shotgun" model for simplification (many simpler sentences) instead of the original sentences improves the chances that the PPI engine will detect a relationship hidden in complex syntax. The precision of the present system is slightly lower than that of the older system because of the exhaustive simplification. This was not the case with the old system where the rules were based on Siddharthan's work and didn't stress as much exhaustiveness of simplification. In situations where precision is much more important than recall, one could use only a subset of the rules that are empirically found to be more precise. The slight loss in precision could also be attributed to the removal of domain-specific features like replacing all gene names with unique identifiers.

**Table 3:** Results for applying protein-interaction extraction system PIE (low precision setting) to sentences from AIMed, as compared to the same to sentences processed with BioSimplify.

|  | Precision | Recall | F-score |
|---|---|---|---|
| Original sentences | 53 | 47 | 50 |
| Sentences simplified by older[11] BioSimplify | 55 | 44 | 49 |
| Sentences simplified by current BioSimplify | 49 | 67 | **57** |

**Table 4:** Results for tight PIE

|  | Precision | Recall | F-score |
|---|---|---|---|
| Original sentences | 46 | 58 | 51 |
| Sentences simplified by older[11] BioSimplify | 51 | 64 | 57 |
| Sentences simplified by current BioSimplify | 46 | 82 | **60** |

Three judges evaluated the precision and recall of the BioSimplify system itself by reading each simplified sentence produced. The evaluation criteria (grammatical correctness) and this post-hoc evaluation using judges follow the same model and rationale described by Siddharthan[4]. We used 404 sentences from AIMed for this evaluation, for an estimated precision of 90%. Since the evaluation for grammatical correctness is done post-hoc by humans, the recall could be overestimated, as it is cognitively difficult to think of all possible grammatically correct sentences. Our judges found less than 1% new simpler sentences that were not produced by the system. The corpus of 4511 biomedical sentences (out of which 2017 contain PPIs) produced from the original 404 is available at https://biosimplify.sourceforge.net. These sentences were artificially created by BioSimplify from AIMed corpus and are manually annotated to indicate the presence or absence of PPI(s).

**Acknowledgments**
We wish to acknowledge our colleagues – Laura, Annie and Robert, native English PhD students with Biology majors, for diligently annotating the hybrid corpus produced by BioSimplify for PPIs. The same team also evaluated the accuracy of the BioSimplify system. Thanks to Kang from Erasmus MC for providing performance statistics about different POS taggers and chunkers. Special thanks to David McClosky, Filip Ginter and Sampo Pyysalo for their kind suggestions regarding the use of parsers.

**References**

1. Harris Z. Distributional Structure. *Word*. 1954;10(2/3):146–62.
2. Quirk R, Greenbaum S, Leech G, Svartvik J. *A comprehensive grammar of the English language*. New York: Oxford Univ Press; 1985.
3. Gee JP. *An introduction to discourse analysis: Theory and method*. London: Routledge; 1999.
4. Siddharthan A. Syntactic Simplification and Text Cohesion. *Research on Language and Computation*. 2006;4(1):77-109.
5. Carroll J, Minnen G, Canning Y, Devlin S, Tait J. Practical simplification of English newspaper text to assist aphasic readers. In: *Workshop on Integrating Artificial Intelligence and Assistive Technology*.; 1998.
6. Vanderwende L, Suzuki H, Brockett C, Nenkova A. Beyond SumBasic: Task-focused summarization with sentence simplification and lexical expansion. *Information processing and management*. 2007;43(6):1606–1618.
7. Jonnalagadda S, Tari L, Hakenberg J, Baral C, Gonzalez G. Towards effective sentence simplification for automatic processing of biomedical text. In: *NAACL HLT*.; 2009.
8. Halliday MA. *An introduction to functional grammar*. London: Arnold; 1985.
9. Hakenberg J, Leaman R, Nguyen V, et al. Efficient extraction of protein-protein interactions from full-text articles. *Accepted by IEEE/ACM TCBB*. 2010.
10. Chandrasekar R, Srinivas B. Automatic induction of rules for text simplification. *Knowledge-Based Systems*. 1997;10:183-190.
11. Jonnalagadda S, Gonzalez G. Sentence Simplification Aids Protein-Protein Interaction Extraction. In: *Languages in Biology and Medicine*.; 2009.
12. Lease M, Charniak E. Parsing Biomedical Literature. In: *International Joint Conference on Natural Langage Processing*.; 2005.
13. McClosky D, Charniak E. Self-Training for Biomedical Parsing. In: *ACL*.; 2008.
14. Ade A, Wright Z, Jagadish H. *The NLP web service [Internet]*. Ann Arbor (MI): National Center for Integrative Biomedical Informatics; 2009. Available at: http://nlp.ncibi.org.
15. Tari L, Tu PH, Hakenberg J, et al. Integrating querying and retrieval for biomedical information extraction. *ICDE*. 2009.
16. Marcus MP, Santorini B, Marcinkiewicz MA. Building a Large Annotated Corpus of English: The Penn Treebank. *Computational Linguistics*. 993;19(2):313-330.
17. Bunescu R, Ge R, Kate RJ, et al. Comparative experiments on learning information extractors for proteins and their interactions. *Artif.Intell.Med.* 2005;33(2):139-155.
18. Kim S, Shin SY, Lee IH, et al. PIE: an online prediction system for protein-protein interactions from text. *Nucleic acids research*. 2008.